\documentclass[conference,a4paper,flushend]{iaria} % (based on IEEEtran.cls)
\usepackage{hyperref}
\usepackage{url}
\title{Stress-Testing the Robustness of LLM-Based Causal Inference}

\author{
    \IEEEauthorblockN{
        Ankitkumar Patel\IEEEauthorrefmark{1}, 
        Jigarkumar Patel\IEEEauthorrefmark{2}, 
        Amit Kumar\IEEEauthorrefmark{3}, 
        Supreetha Sreeram\IEEEauthorrefmark{4}, and 
        Venkatesh Kulkarni\IEEEauthorrefmark{5}
    }
    \IEEEauthorblockA{
        \IEEEauthorrefmark{1}IntelliQuest Venture, USA,
        \IEEEauthorrefmark{2}University of Texas at Dallas, USA,
        \IEEEauthorrefmark{3}Texas A\&M University--Corpus Christi, USA \\ 
        \IEEEauthorrefmark{4}Pondicherry University, India,
        \IEEEauthorrefmark{5}IIT - Guwahati, India. \\
        \small
        \texttt{ankitkumar.patel179@gmail.com},
        \texttt{jigarkumar.patel@utdallas.edu},
        \texttt{akumar3@islander.tamucc.edu}, \\
        \texttt{supreme172051@gmail.com},
        \texttt{venkateshkulkarni2001@gmail.com }
    }
}

\usepackage{tcolorbox}
\tcbuselibrary{breakable}
\usepackage{graphicx}
\usepackage{subcaption}
\usepackage{titlesec}

\begin{document}
\maketitle
\begin{abstract}
Why do some of the world’s most powerful Artificial Intelligence (AI) systems still struggle to reason about cause and effect? Causal discovery, the ability to discern the underlying drivers of a system, is fundamental to scientific inquiry and has traditionally relied on structured data and statistical methods. Recently, Large Language Models (LLMs) have been explored as an alternative paradigm for causal inference, leveraging their broad knowledge and reasoning capabilities. Yet, despite their linguistic fluency, LLMs often rely on surface-level pattern recognition rather than genuine causal logic. While it is known that LLMs stumble in this domain, we lack a precise map of where, why, and how significantly these failures occur. In this paper, we propose a systematic stress-test framework across six critical dimensions: model scale, causal graph complexity, prompting strategy, metadata quality, data quality, and sensitivity to uncertainty. Our findings reveal a significant "optimism bias" in LLMs; models consistently overestimate causal links, leading to low precision across the board. Furthermore, we demonstrate that traditional evaluation metrics can be deceptive: while reasoning-heavy models lead in performance, the highest scores often occur when uncertainty in inference is overlooked. Once uncertainty is integrated into the evaluation, performance drops, providing a more honest reflection of causal reliability. We also find that, while scaling model size improves outcomes, these gains are fragile; performance degrades sharply as the quality of metadata decreases. This work serves as a practical guide to the blind spots of LLMs, offering a clear-eyed assessment of when these tools can be integrated into a causal discovery pipeline and when they are likely to lead researchers astray.

% Code for data and reproducing our experiments is available at \url{https://github.com/aamitssharma07/uncertainty-aware-llm-causal-discovery-repro.git}.

\end{abstract}

\section{Introduction}
Causal discovery, the task of inferring causal structure from observational data, has traditionally relied on statistical algorithms grounded in conditional independence testing and score optimization \cite{glymour2019review}. Recently, interest has grown in whether LLMs can serve as a new kind of causal reasoner by leveraging vast pretraining knowledge rather than numerical analysis.

However, early benchmarks such as CORR2CAUSE~\cite{jin2024large} revealed that LLMs (a) barely exceed random chance, (b) fail to generalize, and (c) collapse under simple surface-level perturbations, such as variable renaming. The authors argue that these models rely on lexical and positional cues rather than genuine causal reasoning. Building on this critique, Wu et al.~\cite{wu2025llm} and the broader causal landscape analysis ~\cite{llm_causal_landscape_2024} showed that embedding LLM outputs as prior knowledge can undermine traditional causal discovery methods, with many reported gains attributable to engineered prompts rather than true inference. In contrast, Darvariu et al.~\cite{darvariu2024largelanguagemodelseffective} suggest that LLMs can serve as effective priors for causal graph discovery, while Ban et al.~\cite{ban2025integratinglargelanguagemodel} demonstrate performance improvements when LLMs are used as heuristic guides within causal pipelines. Notably, the authors convinced that LLMs should play a role of a heuristic for search initialization rather than causal discovery.

Despite these insights, these previous works \cite{jin2024large, wu2025llm} share four significant methodological limitations: (a) evaluations are confined to small-scale networks, (b) metadata is treated as a binary condition (present or absent), (c) the influence of varying degrees of description richness on causal inference is not systematically characterized, and (d) the impact of various prompting strategies on causal discovery is not analyzed.

This paper addresses these gaps by proposing a systematic stress test framework on six critical dimensions; (1) model scale, (2) causal graph complexity, (3) prompting strategy, (4) metadata quality, (5) data quality, and (6) sensitivity to uncertainty. We also present a novel uncertainty-aware metric to evaluate causal inference. Our sensitive analysis across a subset of dimensions reveals that the failure modes identified by  \cite{jin2024large, wu2025llm} do not scale linearly. Instead, specific failure patterns are qualitatively amplified at scale, and strategies effective for small graphs can become counterproductive as complexity grows. Notably, richer metadata improves performance on small networks. These findings provide a comprehensive map of the boundary conditions governing LLM reliability.

The rest of the paper is structured as follows. In Section \ref{problem}, we present the formal problem statement of causal discovery. The proposed framework for multi-dimensional sensitivity is outlined in Section \ref{framework}, and the proposed uncertainty-aware evaluation metric is defined in Section \ref{metric}. Section \ref{eval} covers the performance analysis of state-of-art LLMs under the proposed framework. Finally, our findings and future work are summarized in Section \ref{summary}.

\section{Problem Statement} 
\label{problem}

The primary goal of causal discovery is to recover the latent Directed Acyclic Graph (DAG) $G = (V, E)$ from observed variables $V = \{v_1, \dots, v_n\}$. Accurately identifying these edges is essential for scientific explanation and intervention. While LLMs are increasingly used as heuristic priors in this domain, their reliability depends on interacting boundary conditions that remain poorly understood. We formalize this LLM-based inference as a function: $\hat{E} = \mathcal{F}(M, A, P, Q, D)$, where $M$ denotes a model, $A$ denotes a pair attributes, $P$ denotes a prompt configuration, $Q$ denotes metadata quality, and $D$ denotes data. The fundamental challenge is to treat $\mathcal{F}$ as a stable reasoner, ignoring its extreme sensitivity to perturbations in $M$, $P$, $Q$ and $D$.

\section{Multi-Dimensional Sensitivity Analysis}
\label{framework}

To provide a comprehensive map of LLM reliability, causal discovery can be analyzed across the following six domains. 
\subsection{Graph Scale and Topology Complexity}
A causal graph’s complexity is defined by its node count and edge density. As variables increase, the space of potential causal hypotheses grows combinatorially, creating two compounding challenges for LLMs. First, larger graphs result in crowded contexts, forcing the model to process overwhelming lists of relationships and descriptions. Second, finite context-length constraints risk implicit pruning, where the model deprioritized or truncates critical information. These factors limit the model toward surface heuristics rather than genuine reasoning. Consequently, LLM performance often exhibits sharp failures upon reaching capacity limits rather than degrading smoothly with topological complexity.

\subsection{Model Scale and Pretraining Objectives}
LLM effectiveness in causal discovery depends on non-linear interactions between model scale, training data, and pretraining objectives. This complexity makes it difficult to attribute performance gains solely to model size. Empirically, models optimized for reasoning subject to outperform standard LLMs on structured causal assessments. While next-token prediction models often rely on superficial lexical or frequency-based cues, true causal discovery requires specific inductive biases. Consequently, increasing model capacity only improves causal reasoning when paired with objectives that promote abstraction, counterfactual thinking, and logical consistency.

\subsection{Prompting Strategy and Heuristic Bias}
Prompting strategy centralizes LLM causal behavior, with methods like zero-shot, few-shot, and Chain-of-Thought (CoT) significantly altering performance. While CoT can elicit structured reasoning and uncover implicit assumptions through step-by-step explanation \cite{wei2022chain}, its benefits are scale-dependent. As graph complexity increases, CoT prompts induce heavy context overhead, often leading to verbose but shallow reasoning. This can force models to prioritize narrative coherence over causal validity, amplifying heuristic biases. Consequently, prompting strategies that appear effective for small-scale problems may become counterproductive in large-scale problems.
%, more complex settings due to this strong interaction between strategy, graph scale, and context length.

\subsection{Metadata Quality and Description Richness}
Metadata refers to the semantic labels and textual descriptions of variables. While richer metadata can reduce ambiguity and provide the contextual cues necessary for causal reasoning, but on the other hand, it introduces contextual complexity which can overwhelm the reasoning capacity of LLMs. Consequently, the impact of metadata on causal accuracy depends on a non-linear trade-off between description precision, prompt structure, and the graph scale.

\subsection{Observational Data Precision}
The influence of observational data on LLM-based causal reasoning remains an open question, as interpretations are shaped by how data is represented and contextualized within a prompt rather than by numerical content alone.
In practice, the granularity of observations is critical. While coarsely aggregated data may obscure causal signals, overly fine-grained representations can introduce noise and inflate prompt complexity. Consequently, LLM decision-making varies significantly based on data precision and formatting. This necessitates a careful balance between informational fidelity and contextual clarity in causal discovery tasks.

\subsection{Impact of Evaluation Metrics}
Conventional metrics like precision, recall, and F-score assume a binary framework, conflating uncertainty with error. This obscures the critical distinction between an incorrect assertion and a principled abstention, a limitation especially pronounced in LLMs, which lack classically calibrated probabilities. In our setting, we query LLMs with dual prompts for the existence ($P(e)$) and non-existence ($P(\neg e)$) of causal edges. Because LLMs rely on heuristic reasoning rather than coherent probabilistic inference, they frequently violate basic constraints (e.g., $P(e) + P(\neg e) = 1$). We also observe borderline cases where nearly equivalent probabilities reflect epistemic uncertainty rather than a meaningful causal preference. Treating these as hard decisions artificially inflates false positives and negatives. To better assess the trade-off between decisiveness and reliability, we propose a novel ternary decision framework ($True$, $False$, and $Uncertain$). By categorizing inconsistent or marginal predictions as "Uncertain," we can penalize confident errors more severely than informed abstentions. Our proposed metric is formally defined in the following section.

\section{Uncertainty-Aware Evaluation Theory}
\label{metric}

LLM-assisted causal discovery operates under inherent epistemic uncertainty, which is not adequately captured by traditional binary evaluation metrics such as precision, recall, and F-score. Unlike classical statistical methods that enforce hard decisions, LLMs naturally admit a third outcome, abstention, when confidence is insufficient. To account for this behavior, we adopt a ternary decision framework in which each ordered variable pair has a ground-truth state of \textit{Edge} or \textit{No-Edge}, while the model may predict \textit{Edge}, \textit{No-Edge}, or \textit{Uncertain}. This yields a ternary confusion structure that distinguishes uncertainty over true edges (UE) from uncertainty over non-edges (UN), reflecting their differing implications for causal reasoning and intervention planning.

We incorporate this structure into uncertainty-adjusted precision and recall, defined as
\[
R_{adj} = \frac{TP}{TP + FN + \beta \cdot UE}, \quad
P_{adj} = \frac{TP}{TP + FP + \alpha \cdot UN},
\]
where $\alpha, \beta \in [0,1]$ control the penalty assigned to abstention. These parameters encode epistemic preferences between decisiveness and caution. The resulting F-Adj metric, defined as the harmonic mean of $P_{adj}$ and $R_{adj}$, generalizes the classical F-score by explicitly modeling uncertainty.

The weighting parameters \((\alpha, \beta)\) in the uncertainty-aware evaluation metric encode explicit epistemic policies that regulate the trade-off between decisiveness and caution in causal assessment. Rather than treating uncertainty as uniformly undesirable, these parameters allow evaluators to specify when a model should abstain versus when it should commit to a causal claim. Symmetric settings (\(\alpha = \beta\)) reflect balanced treatment of uncertainty, while asymmetric configurations express task-dependent priorities. For instance, \(\alpha > \beta\) favors edge assertion by penalizing uncertainty over non-edges, whereas \(\alpha < \beta\) prioritizes caution by discouraging uncertain edge claims. Such distinctions are critical in causal discovery, where the relative costs of false positives and false negatives vary across domains. By making these epistemic assumptions explicit, the framework aligns evaluation with downstream objectives and supports principled transitions between exploratory and confirmatory stages of a causal discovery pipeline. Table~\ref{tab:weighting-strategies} summarizes representative weighting strategies and their corresponding epistemic goals.

% \begin{table}[ht]
% \centering
% \caption{Weighting strategies and interpretations.}
% \label{tab:weighting-strategies}
% \small
% \begin{tabular}{|c|c|p{6.5cm}|}
% \hline
% $\alpha$ & $\beta$ & Epistemic Goal \\ \hline
% 0.5 & 0.5 & Balanced and symmetric treatment of uncertainty \\ \hline
% 1.0 & 1.0 & Uncertainty treated as full failure; decisiveness enforced \\ \hline
% 0.8 & 0.4 & Silence preferred over guessing a causal edge \\ \hline
% 0.4 & 0.8 & Guessing a causal edge preferred over silence \\ \hline
% 0.2 & 0.8 & Aggressive discovery; prioritize recall over caution \\ \hline
% 0.8 & 0.2 & Conservative discovery; prioritize precision and abstention \\ \hline
% \end{tabular}
% \end{table}

\begin{table}[t]
\centering
\caption{Representative weighting strategies and their epistemic interpretations.}
\label{tab:weighting-strategies}
\footnotesize
\renewcommand{\arraystretch}{1.12}
\setlength{\tabcolsep}{6pt}
\begin{tabular}{c p{0.68\columnwidth}}
\hline
$(\alpha,\beta)$ & Interpretation \\
\hline
$(0.5, 0.5)$ & Balanced and symmetric treatment of uncertainty. \\
$(1.0, 1.0)$ & Uncertainty treated as full failure; decisiveness is enforced. \\
$(0.8, 0.4)$ & Abstention is preferred over guessing a causal edge. \\
$(0.4, 0.8)$ & Guessing a causal edge is preferred over abstention. \\
$(0.2, 0.8)$ & Aggressive discovery setting that prioritizes recall over caution. \\
$(0.8, 0.2)$ & Conservative discovery setting that prioritizes precision and abstention. \\
\hline
\end{tabular}
\end{table}

\section{Performance Evaluation}
\label{eval}

All experiments were conducted on the Asia network, a well-established benchmark from the Bayesian network literature \cite{lauritzen1988local}. The dataset models a small medical diagnostic system with eight binary variables related to pulmonary diseases and clinical symptoms. Despite its limited scale, the Asia network captures key causal motifs such as confounding and mediation, making it a canonical testbed for controlled causal discovery. The ground-truth structure is provided as a directed acyclic graph (DAG) encoding medically grounded causal relationships, including the effects of travel and smoking on disease outcomes and downstream symptoms, enabling precise evaluation of structural correctness and error types.

%*******With ciations**********
LLMs are evaluated under a zero-shot, instruction-based prompting setting. Specifically, we consider \texttt{GPT-4.1-mini} \cite{openai2025gpt41mini}, \texttt{GPT-4.1} \cite{openai2025gpt41}, \texttt{GPT-4o} \cite{openai2024gpt4o}, \texttt{DeepSeek-R1} \cite{deepseekai2025r1}, \texttt{GPT-5.2} \cite{openai2025gpt52}, and \texttt{gpt-oss-20b} \cite{openai2025gptoss20b}. All models are accessed through a unified API-based evaluation pipeline and receive an identical task description along with the same rule-based instructions for identifying adjacency, spuriousness, and independence between attributes and metadata. This setup minimizes confounding effects from advanced prompt engineering, such as exemplar bias and context overload, and allows performance differences to be attributed more directly to model behavior and metadata quality. Full prompt templates and implementation details are provided in \cite{patel2026}; exploration of more advanced prompting strategies is left for future work.

% LLMs are evaluated using a zero-shot, instruction-based prompting strategy. Specifically, we evaluate \texttt{GPT-4.1-mini}, \texttt{GPT-4.1}, \texttt{GPT-4o}, \texttt{DeepSeek-R1}, \texttt{GPT-5.2}, and \texttt{GPT-OSS-20B}. All models are accessed through a unified API-based evaluation pipeline and are provided with an identical task description and the same rule-based instructions for detecting adjacency, spuriousness, and independence between attributes and metadata. This design reduces potential confounding effects arising from advanced prompt engineering, such as exemplar bias or context overload, thereby enabling a clearer attribution of performance differences to model behavior and metadata quality. Full prompt templates and implementation details are provided in \cite{patel2026}, while the exploration of more advanced prompting strategies is deferred to future work.

% \begin{figure}[ht!]
%     \centering
%     \includegraphics[width=0.8\columnwidth]{imgs/gpt-4o.png}
%     \caption{Adjusted F1 across $\alpha$ and $\beta$.}
%     \label{fig:gpt_4o}
% \end{figure} 

\begin{figure}[t]
    \centering
    \includegraphics[width=\columnwidth]{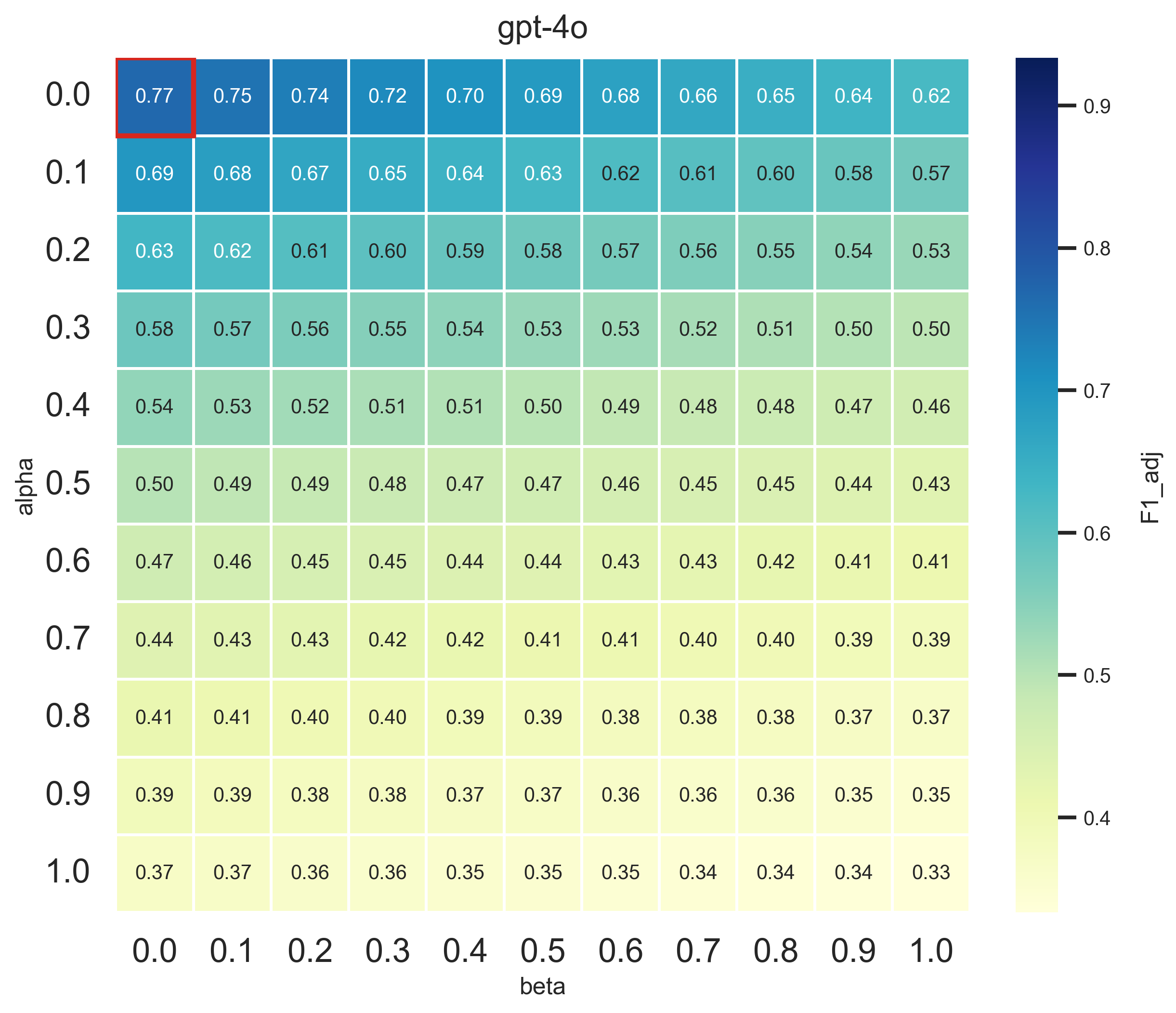}
    \caption{Adjusted F1 of \texttt{gpt-4o} under varying $\alpha$ and $\beta$.}
    \label{fig:gpt_4o}
    \vspace{-4mm}
\end{figure}
To examine the impacts of metadata systematically, we provided metadata with different level of richness, simulating increasing levels of noise and ambiguity. For Asia, metadata is provided at three informativeness levels (L1--L3). At L1, only variable names are provided, introducing maximum noise and forcing reliance on lexical cues. L2 augments names with variable roles and brief descriptions, reducing ambiguity without revealing explicit structure. L3 further explicit causal hints, latent variables, and confounding structure, making portions of the underlying graph approximately recoverable.

% \begin{figure}[ht!]
%     \centering
%     \includegraphics[width=0.8\columnwidth]{imgs/gpt-oss-20b.png}
%     \caption{Adjusted F1 across $\alpha$ and $\beta$.}
%     \label{fig:gpt_oss_20b}
% \end{figure}

\begin{figure}[t]
    \centering
    \includegraphics[width=\columnwidth]{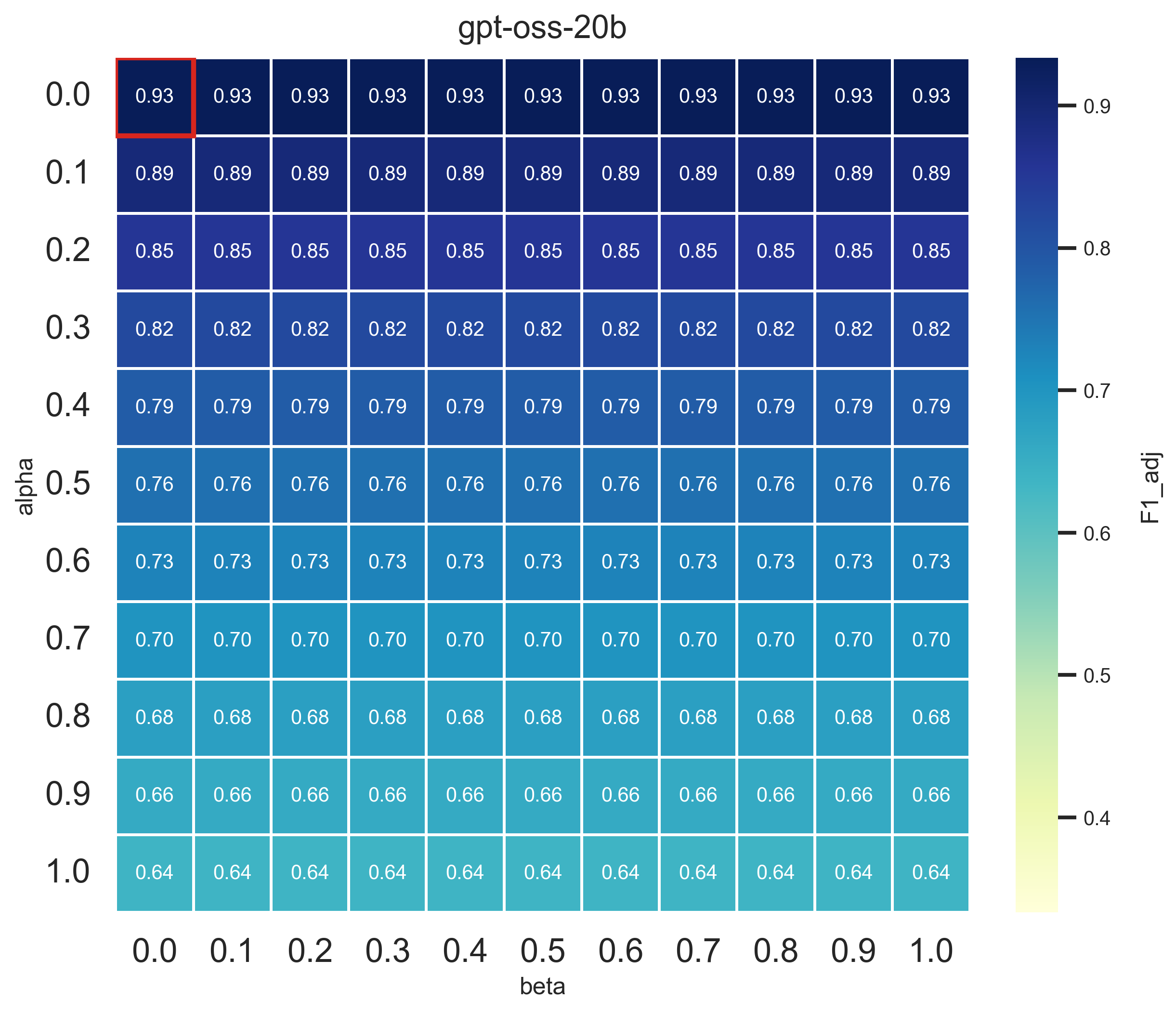}
    \caption{Adjusted F1 of \texttt{gpt-oss-20b} under varying $\alpha$ and $\beta$.}
    \label{fig:gpt_oss_20b}
\end{figure}

This controlled degradation framework enables the isolation of metadata-driven failure modes and distinguishes errors due to limited semantic grounding.

For each metadata variant, each model was queried twice using a simple vanilla prompting strategy: once with an \textit{edge} prompt to estimate the likelihood of a directed causal relation between a variable pair, and once with a \textit{no-edge} prompt to estimate the likelihood of no direct causal relation. In this setting, the LLM is provided only with the task description and metadata. This design choice minimizes confounding effects introduced by advanced prompt engineering, such as exemplar bias or context overload, and enables clearer attribution of performance differences to intrinsic model behavior and metadata quality. We used a fixed temperature of 0.0 and decision threshold of 0.7. For uncertainty-aware evaluation, the penalty parameters $\alpha$ and $\beta$ were varied over $\{0.0, 0.1, \dots, 1.0\}$, and predictions were aggregated into adjusted metrics, including \texttt{F1\_adj}.

\begin{figure}[t]
    \centering
    \includegraphics[width=\columnwidth]{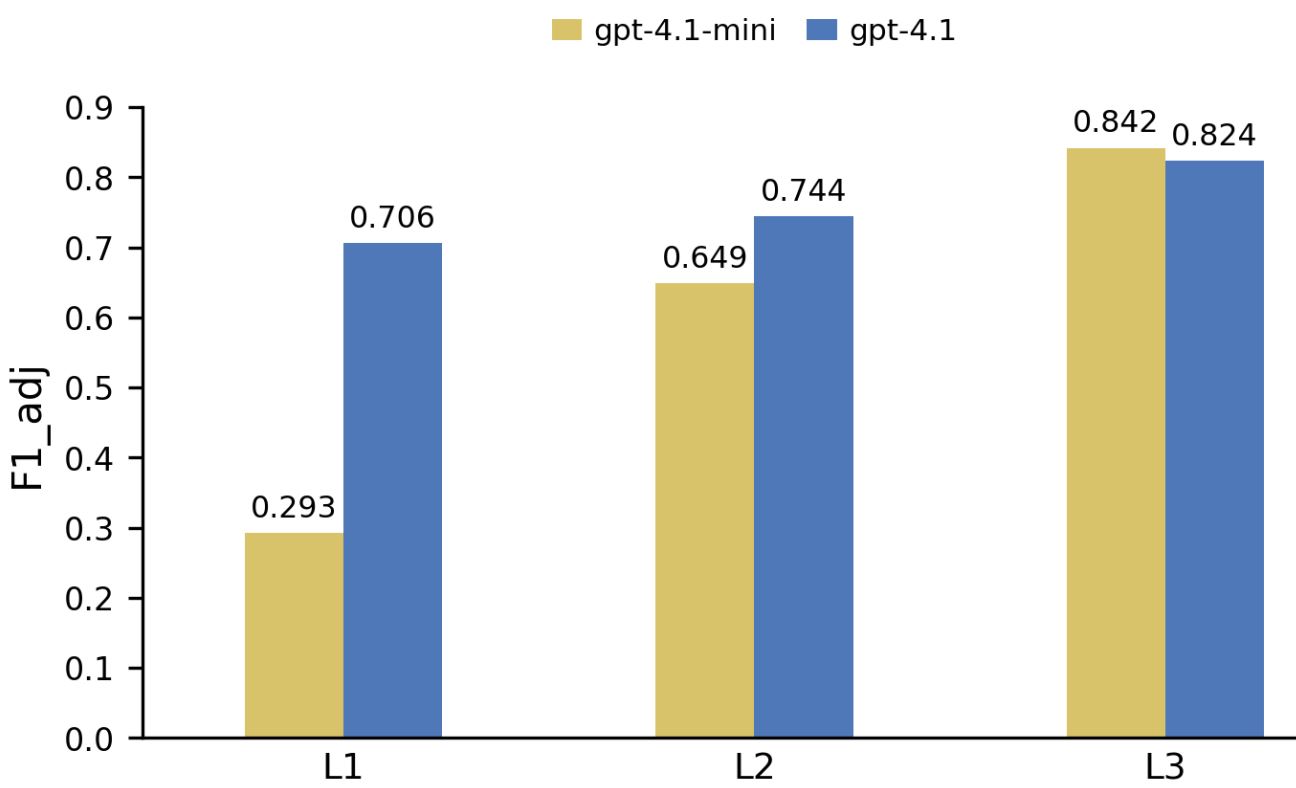}
    \caption{Adjusted F1 as a function of metadata quality on the Asia dataset.}
    \label{fig:small_to_mid_alpha_0.5_beta_0.5}
\end{figure}

% \begin{figure*}[ht!]
%     \centering
%     % Figure A
%     \begin{subfigure}[b]{0.32\textwidth}
%         \centering
%         \includegraphics[width=\linewidth]{imgs/pap_asia.jpg}
%         \caption{F1 adjusted vs. Quality}
%         \label{fig:small_to_mid_alpha_0.5_beta_0.5}
%     \end{subfigure}
%     \hfill
%     % Figure B
%     \begin{subfigure}[b]{0.33\textwidth}
%         \centering
%         \includegraphics[width=\linewidth]{imgs/a2.png}
%         \caption{Metrics vs. Models}
%         \label{fig:radar}
%     \end{subfigure}
%     \hfill
%     % Figure C
%     \begin{subfigure}[b]{0.33\textwidth}
%         \centering
%         \includegraphics[width=\linewidth]{imgs/box_plot.png}
%         \caption{Adjusted F1 across Models}
%         \label{fig:Box_by_model}
%     \end{subfigure}

%     \caption{Comparative evaluation of model performance, metric distributions, and error types.}
%     \label{fig:combined_evaluation}
% \end{figure*}

The heatmaps in Fig. \ref{fig:gpt_4o} and Fig. \ref{fig:gpt_oss_20b} show how model performance ($F1_{adj}$) varies as a function of the uncertainty penalty parameters, $\alpha$ and $\beta$. In all models, the highest scores ($max$) are consistently observed in $(\alpha=0.0, \beta=0.0)$, where uncertainty is effectively ignored. As $\alpha$ and $\beta$ increase, forcing the models to be penalized for ambiguous causal claims, the $F1_{adj}$ scores drop significantly. Among the models, GPT-5.2 and DeepSeek-R1 demonstrate near-perfect performance ($1.00$) maintaining high stability even as the penalty for uncertainty increases. GPT-4o and GPT-OSS-20B show high sensitivity to $\alpha$ and $\beta$.  The score of GPT-4o is degraded by approximately 0.29 as decisiveness is enforced. GPT-OSS-20B exhibits the most dramatic drop, with its performance falling by 0.44 at maximum penalty. This confirms that models often appear more capable than they are by making "guesses" that a binary metric would treat as a hard decision.

In Fig. \ref{fig:small_to_mid_alpha_0.5_beta_0.5}, the performance of gpt-4.1-mini and gpt-4.1 on the Asia dataset was evaluated using the $F1_{adj}$ metric ($\alpha=0.5, \beta=0.5$), which balances incorrect assertions against principled abstentions. The results reveal a non-linear relationship between metadata richness and accuracy. While both models improve as metadata moves from L1 to L3, gpt-4.1-mini shows the most dramatic gain, jumping from 0.293 to 0.842, corresponding to an improvement of approximately 187.4\%. At low richness (L1), the larger gpt-4.1 significantly leads (0.706), indicating stronger internal reasoning when context is sparse. However, as metadata richness increases, the performance gap between the smaller and larger models narrows substantially. At L3, the models converge, and the smaller variant slightly leads (0.842 vs 0.824). In contrast, gpt-4.1 shows a more modest improvement of approximately 16.7\% from L1 to L3, suggesting that richer metadata reduces the performance difference between smaller and larger models.

% \begin{figure}[ht!]
% \centering
%         \includegraphics[width=0.8\columnwidth]{imgs/a2.png}
%         \caption{Metrics vs. Models.}
%         \label{fig:radar}
% \end{figure}
\begin{figure}[t]
    \centering
    \includegraphics[width=0.8\columnwidth]{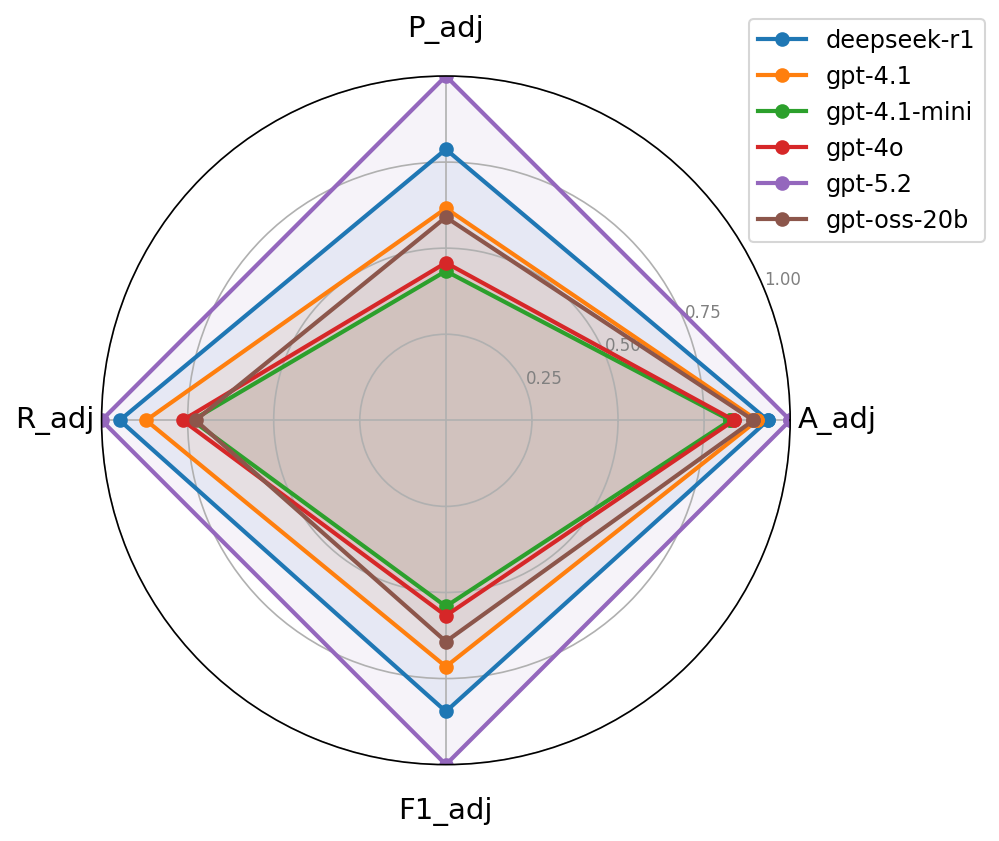}
    \caption{Comparison of evaluation metrics across models.}
    \label{fig:radar}
\end{figure}

The radar chart in Fig. \ref{fig:radar} provides a holistic view of model performance across four adjusted metrics: $P_{adj}$, $R_{adj}$, $A_{adj}$, and $F1_{adj}$ when $\alpha=0.5$ and $\beta=0.5$. Most models exhibit an "elongated" shape toward $R_{adj}$, indicating higher recall than precision. This provides empirical evidence for the "optimism bias", LLMs are inclined to over-identify causal links, leading to a high rate of false positives. GPT-5.2 occupies the outermost perimeter of the chart, showing nearly perfect scores ($1.00$) across all four metrics. DeepSeek-R1 follows closely, though it shows a slightly higher tendency toward recall over precision compared to the top frontier. Smaller models like GPT-4.1-mini and GPT-4o occupy a much smaller area of the radar, with significant dips in $P_{adj}$. This reflects, their tendency to rely on surface-level patterns and metadata hints rather than robust causal logic.

% Added in response to give some concrete examples to show how the reasoning fails for some of the models will help readability and improve the narrative
To improve readability and ground the quantitative results in concrete behavior, we examine representative model outputs. This qualitative inspection reveals clear and intuitive failure patterns that help explain the performance differences in Fig. \ref{fig:radar}. For instance, in the Asia network, several mid- and low-capacity models incorrectly infer a direct causal link from \textit{smoking} to \textit{dyspnoea}, ignoring the mediating disease variables despite metadata that explicitly describes the indirect relationship. In other cases, models assert bidirectional dependencies between clinically unrelated variables, often driven by superficial co‑occurrence cues in the metadata. These errors become more frequent as metadata quality degrades or context becomes crowded. In contrast, GPT‑5.2 more often abstains in such ambiguous situations, which is reflected in its higher and more stable $F1_{adj}$  scores.

% \begin{figure}[ht!]
% \centering
%         \includegraphics[width=1.0\columnwidth]{imgs/box_plot.png}
%         \caption{Adjusted F1 across Models.}
%         \label{fig:Box_by_model}
% \end{figure}

\begin{figure}[t]
    \centering
    \includegraphics[width=\columnwidth]{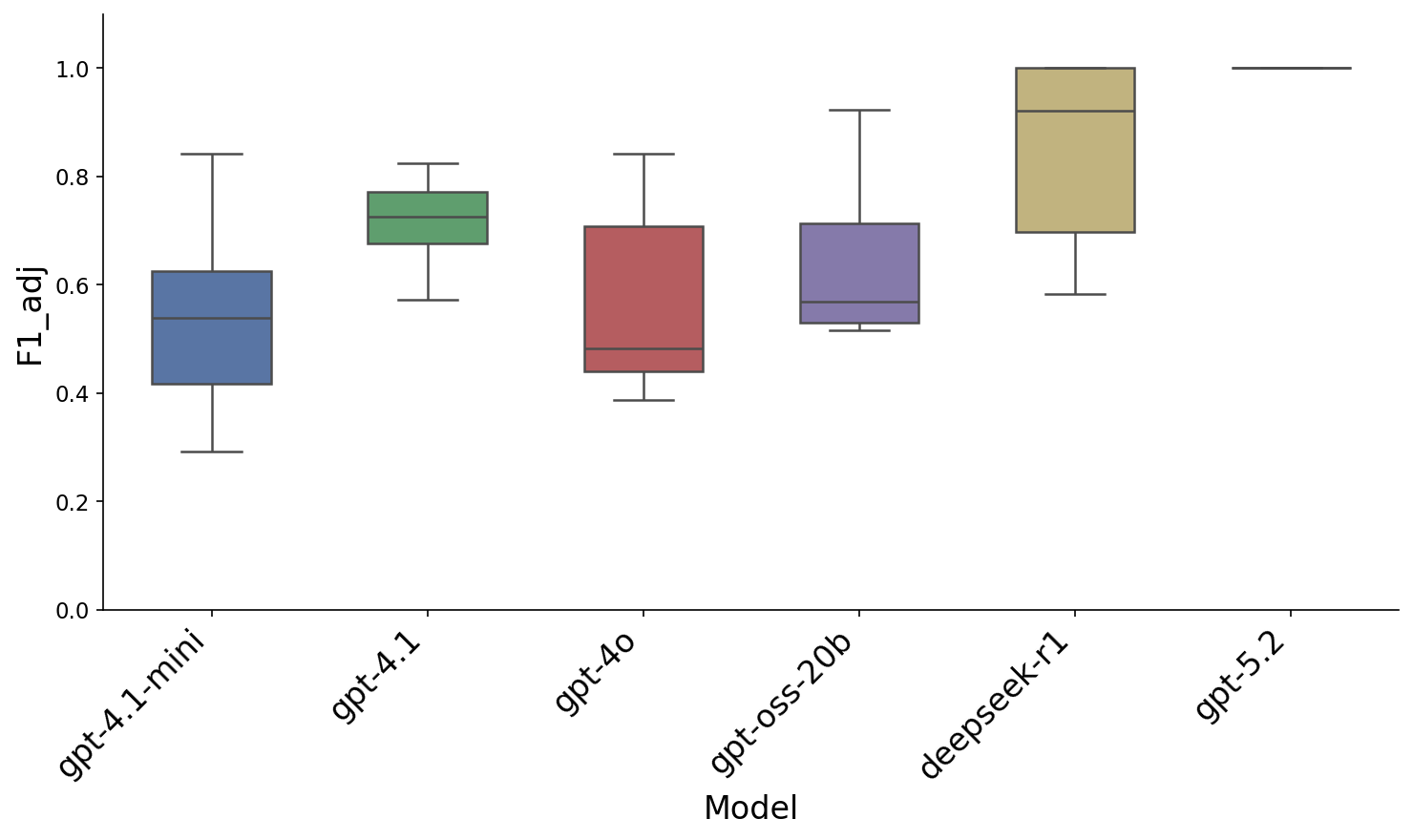}
    \caption{Distribution of adjusted F1 scores across models.}
    \label{fig:box_by_model}
\end{figure}

Figure \ref{fig:box_by_model} captures the variance and median performance of $F1_{adj}$ across different models in a box plot when the input context randomly varies between $L1$ to $L3$ levels. GPT‑5.2, with its higher capacity and enhanced reasoning abilities, achieves substantially higher median $F1_{adj}$ scores and tighter performance distributions than smaller or non‑reasoning models, whereas the remaining models exhibit wide inter‑quartile ranges and multiple outliers, indicating inconsistent application of causal reasoning across varying metadata qualities.  

% New paragraph to address why should someone look at other less capable/cheaper models.
GPT‑5.2 delivers the highest and most consistent $F1_{adj}$ scores, however, it also comes at a substantially higher deployment cost which makes it impractical for all use cases. The box‑plot shows that several smaller or cheaper models, such as GPT‑4.1 and DeepSeek‑R1, achieve reasonably stable performance under favorable conditions, suggesting they can be effective when graph size and metadata quality are controlled. The uncertainty‑aware $F1_{adj}$ metric helps make these tradeoffs explicit, revealing when lower‑cost models are reliable enough and when their variability poses unacceptable risk. This enables practitioners to reserve frontier models for high‑stakes decisions while using less expensive models for exploratory analysis or resource‑constrained settings, rather than defaulting to maximum scale by assumption.

% New paragraph to address discussion about Add a small discussion about how this work relates to GenAI4Cloud or Cloud computing in general.
While high-capacity models typically offer superior reliability and performance, they are not always the optimal choice when operating under stringent cost, latency, and scalability constraints. By explicitly accounting for the trade-offs between these operational metrics and raw model performance, practitioners can make informed decisions when adopting the appropriate models for specific applications. Furthermore, this balanced approach enables more efficient resource provisioning and reservation within generative AI cloud environments.

\section{Conclusion and Future Work}
\label{summary}

This paper presents a systematic stress test of LLMs for causal discovery, revealing a consistent optimism bias in which models overestimate causal relationships. To overcome the limitations of binary evaluation, we introduce the uncertainty-aware \(F1_{adj}\) metric, which shows that many apparent performance gains disappear once epistemic uncertainty is accounted for.
While reasoning-optimized models achieve stronger performance, smaller models like GPT-4.1-mini, narrowing the gap with larger models as metadata richness increases. 
Collectively, these findings provide a more realistic and reliable assessment of LLM capabilities in causal reasoning tasks. Building on these boundary conditions, future work will evaluate how LLM judgments shift when provided with raw observational data alongside variable descriptions. We plan to move beyond zero-shot prompting to study how advanced strategies, such as CoT and multi-agent debate, interact with graph scale. Finally, we will expand evaluations to complex systems exceeding 200 nodes to transform LLMs into robust components of high-stakes causal discovery pipelines.

\printbibliography

\end{document}